\documentclass[conference]{IEEEtran}
\IEEEoverridecommandlockouts
\usepackage{cite}
\usepackage{amsmath,amssymb,amsfonts}
\usepackage{algorithmic}
\usepackage{graphicx}
\usepackage{textcomp}
\usepackage{xcolor}
\usepackage{url}
\usepackage{booktabs} 
\def\BibTeX{{\rm B\kern-.05em{\sc i\kern-.025em b}\kern-.08em
    T\kern-.1667em\lower.7ex\hbox{E}\kern-.125emX}}
\begin{document}

\title{Classification of Diabetic Retinopathy using Pre-Trained Deep Learning Models\\
{\footnotesize \textsuperscript{}}
\thanks{}
}

\author{\IEEEauthorblockN{1\textsuperscript{st} Inas Al-Kamachy}
\IEEEauthorblockA{\textit{} \\
\textit{Karlstad University}\\
Sweden \\
inasalka100@student.kau.se}
\and
\IEEEauthorblockN{2\textsuperscript{nd} Prof. Dr. Reza Hassanpour}
\IEEEauthorblockA{\textit{} \\
\textit{Rotterdam University}\\
Netherland \\
zarer@hr.nl}
\and
\IEEEauthorblockN{3\textsuperscript{rd} Prof. Roya Choupani}
\IEEEauthorblockA{\textit{} \\
\textit{Angelo State University}\\
USA \\
rchoupani@angelo.edu}

}

\maketitle

\begin{abstract}
Diabetic Retinopathy (DR) stands as the leading cause of blindness globally, particularly affecting individuals between the ages of 20 and 70. This paper presents a Computer-Aided Diagnosis (CAD) system designed for the automatic classification of retinal images into five distinct classes: Normal, Mild, Moderate, Severe, and Proliferative Diabetic Retinopathy (PDR). The proposed system leverages Convolutional Neural Networks (CNNs) employing pre-trained deep learning models. Through the application of fine-tuning techniques, our model is trained on fundus images of diabetic retinopathy with resolutions of 350x350x3 and 224x224x3. Experimental results obtained on the Kaggle platform, utilizing resources comprising 4 CPUs, 17 GB RAM, and 1 GB Disk, demonstrate the efficacy of our approach. The achieved Area Under the Curve (AUC) values for CNN, MobileNet, VGG-16, InceptionV3, and InceptionResNetV2 models are 0.50, 0.70, 0.53, 0.63, and 0.69, respectively.

\end{abstract}


\section{Introduction}
Diabetic Retinopathy (DR) is considered the first cause of blindness in the world, and the predictions estimate that the number of affected cases will be more than 370 million patients by 2030. The main affecting factor of DR is the clogging of the eye blood vessels, which causes eye reactions in two ways:
The first reaction is creating new blood vessels above the main area (vitreous) which must be clear to allow the light to reach the most sensitive part of the eyes (retina) by passing through the cornea, pupil, and lens. The main function of the retina is to convert the light into impulses that are transmitted through the optic nerve to the brain where the interpretation occurs to help us see and understand scenes.
The second reaction is blood leakage through the blood vessels that will influence and harm the retina specifically in the core part (macula), which is considered essential for detailed vision. The severity levels of diabetic retinopathy have been classified into five classes as shown in Fig 1.
\begin{figure}
  \includegraphics[width=\linewidth]{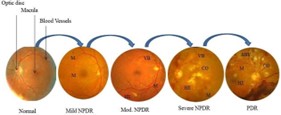}
  \caption{Diabetic Retinopathy Stages}
  \label{fig: Diabetic Retinopathy Stages}
\end{figure}
In this study, we developed a machine learning (ML) application using Convolution Neural Network and four types of pre-trained deep learning models (VGG16, MobileNet, InceptionV3, InceptionResNetV2) for multi-categorical classification of DR. In our experiments, we used 1000 color fundus images from KAGGLE diabetic retinopathy dataset, with 350x350 and 224x224 resolutions. And using several image-augmentation before these images were fed to the pre-trained models.

\section{PREVIOUS WORK}

Shorav Suriyal et al[1] used a dataset of 16,798 color fundus images from Kaggle and classified them into two classes (with DR, without DR) by building a model which can be used on an Android platform on a mobile device. They applied two main phases: Image pre-processing, to remove noise using "Box Blur" and fit all images into 256x256 pixels, and building the model, which was inspired by MobileNet and used transfer learning with 25 convolutional layers. The main aspect of this network was splitting the network "Depth-wise," which used only one filter, and "Pointwise," which used the linear activation function. The authors reported a final accuracy of 73.3\%.

Arkadiusz Kwasigroch et al[2] introduced research that used a dataset from eyepaces.com, including about 88,000 color fundus images with 1500x1500 resolutions, to classify them into five stages of DR.

To normalize the image sizes all training images underwent under-sampling or over-sampling stages. Next, they split the data set into 3 parts (training images with 3500, validation images with 1000, testing images with 1000). 
Their method consists of two main phases: Image pre-processing:
\begin{itemize}
    \item Resizes all images to a fixed size ($224 \times 224$ pixels) and fixes all eye's radius.
    \item Modifies image scale between [0, 1].
    \item Applies image normalization with means equal to zero, and the variance equal to one.
    \item Applies local average subtraction operation.
    \item Confirms data augmentation technique.
    \item Classification phase: This phase uses a transfer-learning technique that is based on the VGG-D model.
\end{itemize}

The model was built using a vast data set by changing the last layer of the network and using the same parameters and weight and bias values that were used in a similar task.
The model incorporates five convolution layers having ReLU as activation functions, where each one is followed by max or average pooling layers, a dropout layer, and a fully connected layer. The final layer uses a non-linearity sigmoid as an activation function with 4 neurons corresponding to the number of classes.
In their study, they used a special technique to label the classes as follows:
([0 0 0]t, [1 0 0]t, [1 1 0]t, [1 1 1]t). 

Using 200 images per class with kappa-score as the accuracy metric the final accuracy was 81.7

Xiaoliang Wang et al [3] used 166 color fundus images with high quality from KAGGLE to classify them into DR stages (5 classes).
The images were noisy, and required size normalization before putting them into three different resolutions to be used as input to three CNN architecture types:

\begin{itemize}
    \item \textbf{AlexNet:} Resize the images to $227 \times 227$ pixels.
    \item \textbf{VGG-16:} Resize the images to $224 \times 224$ pixels.
    \item \textbf{Inception-V3:} Resize the images to $299 \times 299$ pixels.
\end{itemize}

Using a transfer learning technique to make use of the parameters of weight and bias which had been obtained using a huge number of images made them reusable for similar tasks.
The authors used 5-fold cross-validation to split the source images into training images, test images, and validation images. The final accuracy was as follows:
\textbf{\begin{itemize}
    \item Alex Net: 37.43\%.
    \item VGG-16: 50.03\%.
    \item InceptionNet-V3: 63.23\%.
\end{itemize}}

Inception Net had high accuracy due to using a fine-tuned SGDM optimizer to reduce loss function. 
The main weak point in their work is the resizing operation to the original image that causes loss of some features of the image, and the small number of images that had been used compared to the available number.
Safaraz Masood et al [4] used image pre-processing and transfer learning techniques (inception V3) to classify DR into five classes (No DR, Mild, Moderate, Severe, PDR) using a dataset that had been arranged by Eye-Paces on KAGGLE, the total number of images was around 800 for each class except PDR where it was 708 Images).
Below is the main image preprocessing operation:
\begin{itemize}
    \item Reduce the image size radius to 200 pixels.
    \item Highlight the important features using the blue function.
    \item Delete the boundary of the images.
\end{itemize}

By using the transfer learning technique with the same value of weights and biases that have been obtained from the training of different data sets they make use of these values to perform training on the data set of DR.
By using the same architecture of ImageNet (Inception V3), and making Cross-Entropy as a loss function, they conducted 4 experiments, by increasing the number of images to 800 images at each class and radius size to 500 pixels the accuracy was improved.
The final test accuracy is 48.2\% Ardianto et al [5] built a network named "Deep-DR-Net" to classify the input fundus images into three stages (Normal, Mild NPDR, severe NPDR), and reduce the size to 320x240 and 237 images as training dataset using three-phase:

\begin{enumerate}
    \item First: Construct a feature map at the beginning of the network by using both the max-pooling layer and convolution layer to obtain 16 feature maps as input to the next stage of the network.
\end{enumerate}

\begin{enumerate}
    \item Second: Build the coder using a set of convolution blocks in the position that is influenced by CNN architectures (ResNet and InceptionNet), using five cascaded convolution layers where ReLU and batch normalization were between the blocks, with a max-pooling and dropout layers.
\end{enumerate}

\begin{enumerate}
    \item Third: Use softmax for classification after the second phase. By using the FINdERS dataset with 315 fundus images together with "Deep-DR-Net" and SGD (stochastic gradient descent), the final accuracy was 60.82\%.
\end{enumerate}
 
\section{The Data Set}
we used a dataset from the KAGGLE competition which contains 35,126 color fundus images taken from different positions of eyes (left, right), the images were scaled between [0-4] to correspond to the five of DR disease (Normal, Mild, Moderate, Severe, PDR), 

\section{PROPOSED WORK AND METHODOLOGY}
The proposed method was implemented in KAGGLE KERNEL using Python programming languages. We used 1000 color fundus images of the KAGGLE dataset to carry out this research.

\subsection{Image Preprocessing}\label{AA}
We resize the images to resolutions of 350x350x3 and 224x224x3. Next, we divide them into training, validation, and testing datasets, allocating 80\%, 10\%, and 10\% respectively. Then, we standardize the training set $(x_{\text{train}})$ and $(y_{\text{train}})$ by dividing each by the standard deviation (255 in RGB images, which represents the maximum value of a pixel channel), ensuring that each image value falls within the interval $[0, 1]$.
We specify five classes corresponding to the levels of DR and use one-hot encoded vectors for $(y_{\text{train}})$ and $(y_{\text{test}})$ as follows: $[1,0,0,0,0,0,0]$ refers to the Normal class, $[0,1,0,0,0]$ refers to the Moderate class, and so on.

\subsection{Data Augmentation}
To reduce overfitting and increase the number of images to 2000 as well as the accuracy of the model we used image augmentation as an important operation to be applied to both training and testing datasets, using different geometric transforms (shift, flip, zoom, channel shift, rotate, rescale, etc.)

\subsection{Convolutional Neural Network}
We constructed a Convolutional Neural Network (CNN), widely regarded as state-of-the-art for various image classification tasks, for diabetic retinopathy (DR) image classification. Our proposed CNN architecture, illustrated in Figure 3, consists of the following layers:

\begin{itemize}
    \item Convolutional Layer: Used to extract features from input images comprising pixels. We employ 16 kernels of size 3x3 for this purpose.
    \item Batch-Normalization Layer: Normalizes the output of the preceding layer across the batch size.
    \item Max Pooling Layer: Reduces the dimension of the output image using a 2x2 window.
    \item Dropout Layer: Mitigates overfitting by randomly dropping out some neurons during training.
    \item Flatten Layer: Converts the output to a one-dimensional (linear array) to pass it to the next layer.
    \item Dense Layer: A fully connected layer utilized for final classification with 5 classes. It is followed by the non-linear activation function 'SoftMax'.
\end{itemize}

\subsection{Fine Tune}
To avoid time-consuming computational tasks, we make use of the weight parameter of the model that was trained in ImageNet which is around (1.2 M). In addition, we use small datasets which are from different domains compared to ImageNet. Furthermore, we make the original model as an automatic feature extractor, while freezing some numbers of layers, and added several layers using fine-tune operation, which fine-tunes the weight in the base model to be used in the pre-trained model by freezing lower layer to solve the overfitting problem and adding different layers at the bottom. These earlier layers extract uncomplicated features, while deeper layers extract complicated features and learn a high level of features of the whole image. The input image was resized to be 350x350x3 for (VGG16, InceptionV3, InceptionResNetV2), while MobileNet used 224x224x3 as the size of the input image. The distribution of our classes is imbalanced, so we used a batch Normalization layer to convert the distribution of input images between the interval [-1, 1] or make the mean equal to 0 while setting the standard deviation to one.
In this research four types of pre-trained deep learning models were used to classify diabetic retinopathy into five classes corresponding to five levels of the disease (Normal, Mild, Moderate, Severe, PDR), they are (VGG16, MobileNet, InceptionV3, InceptionResNetV2) each one of them was trained using ImageNet as the dataset.

\subsection{VGG16}
The introduced multilayer model used in different deep learning architectures was VGG Net (Visual Geometry Group), and the only pre-processing image has subtracted the mean of (RGB), VGG Net contained four blocks where each one consisting of a stack of following layers: - The convolution layer had a small receptive field of (3x3) and (1x1) filters, and the stride equal to one. Max pooling layer which had (2x2) window and stride equal to two. With 3 fully connected layers, two contained 4029 neurons, and the third contained 1000 neurons. Where the number of convolution layer and Maxpooling layer were respectively (13, 5). The main characteristic of VGG is: - Uses small receptive fields which decrease the computation task and increase the operation of training. Used blocks Ignored normalization layer which leads to an increase in the computational task and cost memory size. They used 'Relu' as the activation function in hidden layers and 'SoftMax' in the output layer with 1000 neurons. The input image was 224x224x3. Finally, the VGG Net indicates that the architecture with deeper layers shows good performance accuracy.

\subsection{Mobile Net}
With 28 layers and by using the same image preprocessing we apply MobileNet as a pre-trained model using the same hyper-parameter and convert the last layer with a dense layer with 5 classes. This model is used for mobile vision applications and embedded systems because of these properties: - The size of the model is small compared with other models. - Computational operations are faster due to depth-wise separable convolution. MobileNet doesn't have a fully connected layer and it uses a kernel with (1x1x3), and the main aspect of MobileNet is depth-wise separable convolutions in two levels: Depth-wise Convolution which is considered as filter level. Pointwise Convolution considers a combination level. The first level is used to reduce computational operation by using one signal channel of input color image instead of 3 channels.

\subsection{InceptionV3}
The main idea of the Inception network was instead of choosing a specific convolution layer with a specific filter size, the Inception network could be concatenated more than the convolution layer and max-pooling layer with different filter sizes and then stack them.

in one block which is named the inception block. That will reduce the number of parameters as well as computational tasks.
The inception network has been developed in many ways: in Inception-v1 which combines (filter, convolution layer, and max-pooling layer) using inception blocks and average pooling instead of a fully connected layer, on the other hand, Inception- v2 introduces batch normalization.
Where Inceptionv3 [6] won the competition in 2015 of ImageNet dataset and achieved part 1, it introduces two aspects: Factorization method in the third stage of inception network architecture which reduces the parameters without any impact on the performance of the network Regularization method It applies the idea about shrinking the resolution and increasing the number of channels by using a variety of shapes of the convolutional layer as shown in Figure 4, which allows making the network deeper without any impact on the performance of the network, where deeper and narrow are considered as the best architecture for the network. In this regard, several changes have been made in the nine-stage of inception architecture (inception blocks) as follows:
Using 3x3 convolution Instead of 5x5
Using 1x7 and 7x1 convolution instead of 5x5 Using 1x3 and 3x1 convolution instead of 3x3

\subsection{InceptionResNetV2}
In 2015 a research report [7] described a new method that allowed building a very deep neural network with thousands of layers avoiding vanishing and exploding gradient problems that lead to degradation.

accuracy of the network. The model by using several residual blocks and stacking them produces a residual network, where each block consists of a "skip connection/shortcut", which takes the output of the layer and converts it into the input to another deeper layer and skips some hidden layers (almost two hidden layers). The combination of two powerful deep neural networks (Inception and Residual) networks will produce a new network named InceptionresNetV2. The main strong side of the inception network is reducing the computational cost, model size, and dimensionality. While in residual networks it is possible to use deeper layers without any impact on the accuracy of the network as well as speed up the training process of the model by using residual block which contained residual connection the depth was equal to 572.

\subsection{Flask}
Flask is considered a micro-framework of Python, it's the lightest and shortest way to build a small web application using a lot of code behind it. After we choose the model that has more accuracy, we build our web application using (Python \& Flask) which can access our model using an HTTP client by uploading DR images and sending a request to the HTTP server which responds to the classified label of the image by indicating the probability of distribution, so we can specify to which class it belongs.

\subsection{Measure Metrics}
There are two main metrics used to measure the performance of the model: Threshold metrics and rank metrics as our dataset had an unbalanced class distribution as shown in Figure 21, we will use AUC (Area under Curve ROC) which is considered as rank metrics and a good measure of performance for the unbalanced dataset. The Roc Curve denotes to receiver of the operating characteristic curve, and has two main aspects [9]:
\begin{itemize}
    \item True Positive Rate (TPR)
    \item False Positive Rate (FPR)
\end{itemize}

Sensitivity/TPR = TP/(TP+FN)	(1)

Specificity/FPR = FP/(FP+TN)	(2)

Accuracy=(TP+TN)/(TP+FP+FN+TN)	(3)

Where:
\begin{itemize}
    \item TPR: True Positive Rate
    \item FPR: False Positive Rate
    \item TP: True Positive value, the label resides to the specific class and classifies correctly.
    \item TN: True Negative value, the label resides to the specific class, and classify wrongly
    \item FP: False Positive value, the label does not reside to the specific class, and classify correctly.
    \item FN: False Negative value, the label does not reside in the specific class and classifies wrongly.
\end{itemize}

Our Convolution neural network which was built from scratch using different layers shows lower performance and lower AUC value which was 0.50 at 16 epochs and the highest loss value at 5.318, which is considered the largest value compared to other pre-trained deep learning models, as shown in Table 2 CNN model shows overfitting between training and validation dataset. On the other hand, using four types of the pre-trained model where each one was fine-tuned by adding several layers and replacing the final layer with a fully connected layer containing five neurons corresponding to the number of DR levels shows a better result. As shown in Table 1 VGG16 shows a lower value of AUC than MobileNet, while InceptionResNetV2 shows the largest value of AUC with 0.69 and lower loss which was equal to 2.191, despite the small number of epochs which was 28 compared to the number of epochs in InceptionV3 which was 43 with loss equal to 3.262.

\begin{table}[htbp]
  \centering
  \caption{Model Comparison}
  \label{tab:model_comparison}
  \begin{tabular}{lcccc}
    \toprule
    Model & ACC test data & AUC & Loss & Early Stop \\
    \midrule
    CNN & 67\% & 0.50 & 5.318 & 16 Epoch \\
    Mobile-Net & 66\% & 0.70 & 3.113 & 65 Epoch \\
    VGG\_16 & 57\% & 0.53 & 4.727 & 85 Epoch \\
    InceptionV3 & 68\% & 0.63 & 3.262 & 43 Epoch \\
    InceptionResNetV2 & 68\% & 0.69 & 2.191 & 28 Epoch \\
    \bottomrule
  \end{tabular}
\end{table}

\section{Conclusions and Future Work}

As indicated by experimental results, our model shows a different performance that is because of the different depth and architecture forms of each one by using CNN build from scratch to train our dataset we obtained less performance due to the small depth as well as the limited number of data that was trained with, which is inefficient as well as time-consuming, while fine-tuned the pre-trained model that was trained on ImageNet as a dataset, and used again for different domains in different tasks by making several changes provide better results.

Fine-tuned pre-trained deep learning model used for classification of the KAGGLE dataset of diabetic retinopathy disease shows better results due to the depth and width of the network as well as the huge number of the training dataset (ImageNet).

The main aspect of MobileNet was using Depth separable Convolution, which led to a shrink the weight size and a decrease in the model size with (88) layers as depth size, producing a better result using 224x224x3 image size than VGG16 which uses 350x350x3 image size produce the lower value of AUC which was 0.53 although of large numbers of epochs which were 85 epochs, due to the depth of model which was (23) layers, which considered quite small as well as the number of epochs which was 29 compared to 65 epochs using MobileNet.

On the other hand, both InceptionV3 and InceptionResNetV2 which had state-of-art computer vision and image classification tasks, show convergence of their result despite variation between the epochs, the ability to shrink the resolution as well as the ability to go deeper and increase the numbers of layers with (159) as a depth of InceptionV3, and (572) as a depth of InceptionResNetV2, and all that staff shows powerfully in the medical task using KAGGLE images dataset of diabetic retinopathy disease.

By training with KAGGLE images using a pre-trained InceptionV3 model, we made more improvements to AUC value to be at (0.63) instead of (0.59) in [8] which used the same model (InceptionV3) after nine epochs with 512x512x3 image resolution and early stop technique, additionally, our accuracy of the testing dataset which was equal to 68

The main operations in Inception-V3 are as follows:

- Image augmentation technique which increases the number of images to be (2000) images.
- Layer-wise fine-tuning pre-trained model by freezing the lower layer level and part of the higher layer level and adding several layers as shown in Figure 4, using a 350x350x3 image resolution.
- Using Adam as an optimizer
- After 43 epochs using the early stopping method, the AUC was 0.63

The improvement in our research was due to increasing the number of images to be (2000) images and split to be (1620) training images, (200) testing images, and (180) validation images as well as the ordering of layers which were added at the end of the model was quite helpful using powerful layer convolutional for feature extraction and normalizing the output of the pre-trained model and using it as input to next layer (convolutional with 1x1 kernel size

and 128 neurons and two blocks of (dropout (0.5), dense) to reduce the overfitting of our model.

Using image augmentation as well as increasing the number of training images can solve overfitting problems.

Diabetic retinopathy images are considered medical images which contain more details with noise and variation, and it is different domains form the ImageNet dataset which contains 10 million labeled images, and 1000 categories with 469x387 resolutions.

Finally, InceptionResNetV2 shows the best performance for medical images using the KAGGLE dataset of diabetic retinopathy disease.

Our future works will include the following:

- Using Google Collaboratory which considers a Jupiter notebook environment that enables writing and running our code in our browser using thousands of images using GPU. Setting the resolution of images to 512x512 or more increases the performance of the model and makes specific details in DR clear.
- Using the Flask program to build a diabetic retinopathy application on a mobile device.
- Making connections between the model and the mobile camera, by taking the picture and utilizing the model to make a diagnosis about the diabetic retinopathy level.

Using these steps anyone who suffers from diabetic disease will be able to check his eyes and see whether there is an infection or not and the progress at which levels of the diabetic retinopathy disease like sugar measurement device or blood pressure device.

\end{document}